%% file: main.tex
\documentclass[10pt,twocolumn,letterpaper]{article}

\usepackage[pagenumbers]{wacv} % To force page numbers, e.g. for an arXiv version


\definecolor{wacvblue}{rgb}{0.21,0.49,0.74}
\usepackage[pagebackref,breaklinks,colorlinks,allcolors=wacvblue]{hyperref}
\usepackage[accsupp]{axessibility}  % Improves PDF readability for those with disabilities.

\def\wacvPaperID{1897} % *** Enter the WACV Paper ID here
\def\confName{WACV}
\def\confYear{2027}

\title{Open-World Scene Deocclusion}

\author{
K.M.S.J. Kumarasinghe$^{1}$ \quad
Neil A. Dodgson$^{1}$ \quad
Fang-Lue Zhang$^{2}$\\[4pt]
$^{1}$Victoria University of Wellington
\quad
$^{2}$University of New South Wales
}

\begin{document}

\twocolumn[{

\maketitle

\begin{center}
\includegraphics[width=\textwidth]{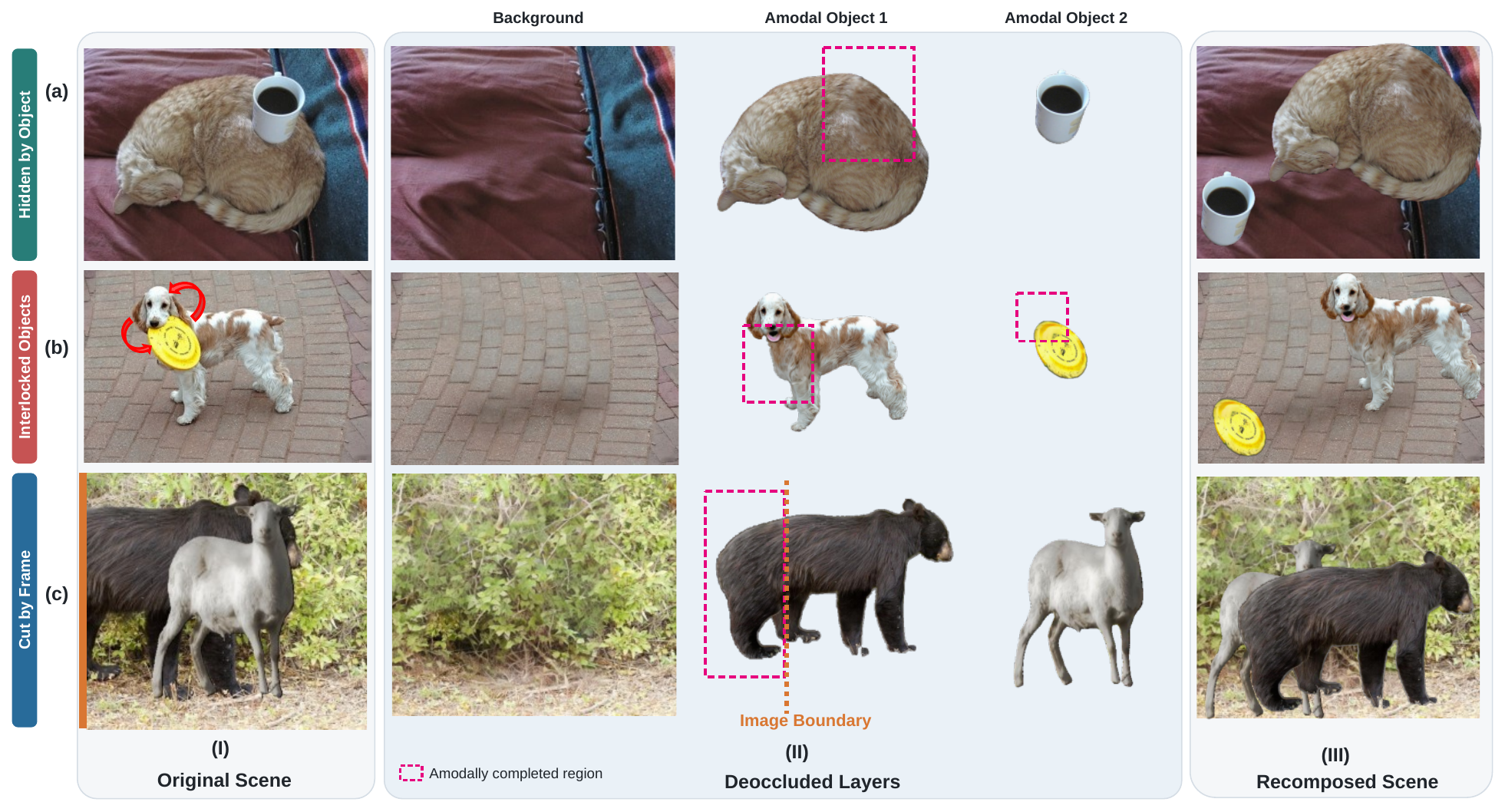}

\vspace{2mm}

\refstepcounter{figure}
\label{fig:teaser}

\parbox{0.95\textwidth}{
\small
\textbf{Figure \thefigure.}
Deocclusion across three scene types: (a) an object hidden behind another, (b) interlocked objects that mutually occlude (red arrows mark the cycle), and (c) an object cut by the image frame (orange line marks the original boundary). Dotted boxes highlight recovered regions. Columns: (I) original scene, (II) deoccluded layers, (III) recomposed scene.}

\end{center}

\vspace{4mm}

}]
\input{sec/0_abstract}

\input{sec/1_intro}

\input{sec/2_relatedwork}
\input{sec/3_method}

\input{sec/4_experiments}

\input{sec/5_conclusion}
{
    \small
    \bibliographystyle{ieeenat_fullname}    
    \bibliography{main}
}

\end{document}

%% file: sec/0_abstract.tex
\begin{abstract}
  Recovering the complete appearance of occluded objects from a single image is a fundamental problem for scene understanding and image editing. Real-world scenes are challenging due to clutter, complex spatial occlusions, and diverse object categories. Existing methods often struggle when occlusion involves background surfaces or objects that are not explicitly detected, and fail to resolve the cyclic occlusion dependencies that frequently occur in natural scenes. We introduce a training-free, open-world scene deocclusion framework that recovers the full amodal appearance of detected or user-selected objects from a single RGB image, decomposing the scene into depth-ordered RGBA layers that are geometrically plausible, semantically faithful, and visually consistent with the visible scene. Extensive experiments show that our approach improves scene-level deocclusion, producing coherent and realistic reconstructions in real-world settings. Code is available on the project page: \url{https://sashjk.github.io/owsd/}
\end{abstract}

%% file: sec/1_intro.tex
\section{Introduction}
\label{sec:intro}

Real-world images are rarely uncluttered \cite{rosenholtz2007measuring}. Objects overlap and occlude, revealing only fragments of their true extent. Humans effortlessly infer the complete shape and appearance of partially hidden objects by mentally ``filling in'' what lies behind. This ability, often known as amodal completion~\cite{restle1981classical}, supports inference beyond what is directly visible \cite{LIER1999203}. Replicating it computationally remains an open challenge: repositioning or removing a partially occluded object requires its full structure rather than only the visible region; otherwise, edits produce implausible geometry and visual inconsistencies.

We define open-world scene deocclusion as recovering the full appearance of all partially occluded objects from a single RGB image. It produces a layer-decomposed representation with geometrically plausible, semantically faithful, and visually consistent completions. In open-world settings, occlusions are layered: an object may be sequentially occluded by several others, including regions treated as background. This demands scene-level occlusion reasoning rather than isolated object completion, together with strong visual priors that generalize across diverse categories.

Although recent large vision-language models (LVLMs) show strong semantic understanding, they are ill-suited to this task, which requires depth aware geometric reasoning and pixel level synthesis beyond their discriminative design  \cite{rahmanzadehgervi2024vision, pothiraj2025capture, watahiki2025bridging, ho2025review}. Among specialized methods, mask based pipelines predict an amodal mask and then synthesize pixels, making them prone to error propagation  \cite{zhan2020self}; diffusion based methods instead generate missing regions directly, achieving stronger realism, but differ in how they condition the process  \cite{ozguroglu2024pix2gestalt,xu2024amodal,ao2025open}. When the goal is full scene deocclusion (rather than just completing a single object), the current state of the art exhibits two failure modes: \textbf{(1)}~Occluding surfaces not segmented during detection are either ignored  \cite{zhan2020self,ozguroglu2024pix2gestalt,10.1145/3641519.3657409,xu2024amodal} or partitioned without depth validation \cite{ao2025open}. As a result, hidden regions may be missed, and the inpainting mask can exclude true foreground occluder pixels. Prior work \cite{ao2025open} identifies background occluders via morphological boundary expansion. However, it does not verify that these regions are physically in front of the target object, leading to false positives from depth-incompatible but spatially adjacent surfaces.
\textbf{(2)}~Real-world scenes frequently exhibit occlusion cycles. Because existing frameworks rely on naïve topological sorts or simple depth-from-camera heuristics, they cannot resolve these cycles, leading to physical contradictions and jagged intersections in the resulting layers.

We address both challenges with a unified, training-free framework for open-world layered scene deocclusion. Given a single RGB image, our framework decomposes the scene into depth-consistent layers and reconstructs the complete amodal appearances of both detected and user-specified objects. It formulates deocclusion as structured inference, jointly reasoning about occlusion relationships and depth to produce a consistent, cycle-free layer ordering. The result is an interpretable decomposition into RGBA layers that preserves physical plausibility. We benchmark our framework against three recent methods \cite{ao2025open,xu2024amodal,ozguroglu2024pix2gestalt} following standard evaluation protocols.

\noindent In summary, our contributions are:
\begin{itemize}
\renewcommand{\labelitemi}{--}
  \item \textbf{A novel framework for layered open-world scene deocclusion.} We formalize recovering the amodal appearance of detected and user-specified objects as a layered, per-object RGBA decomposition for downstream tasks.
  \item \textbf{Depth-validated background occluder detection.} We combine depth cues with morphological analysis to identify background occluders, enabling accurate completion and a consistent background without semantic labels.
  \item \textbf{Cycle-safe layer assignment.} We introduce a depth-aware strategy that produces valid layer ordering and consistent completion under cyclic occlusions, where standard topological sorting fails.
  \item \textbf{Empirical advancement.} We demonstrate consistent improvements over state-of-the-art methods under standard evaluation protocols.
\end{itemize}

%% file: sec/2_relatedwork.tex
\begin{figure*}[!ht]
    \centering
    \includegraphics[width=\textwidth]{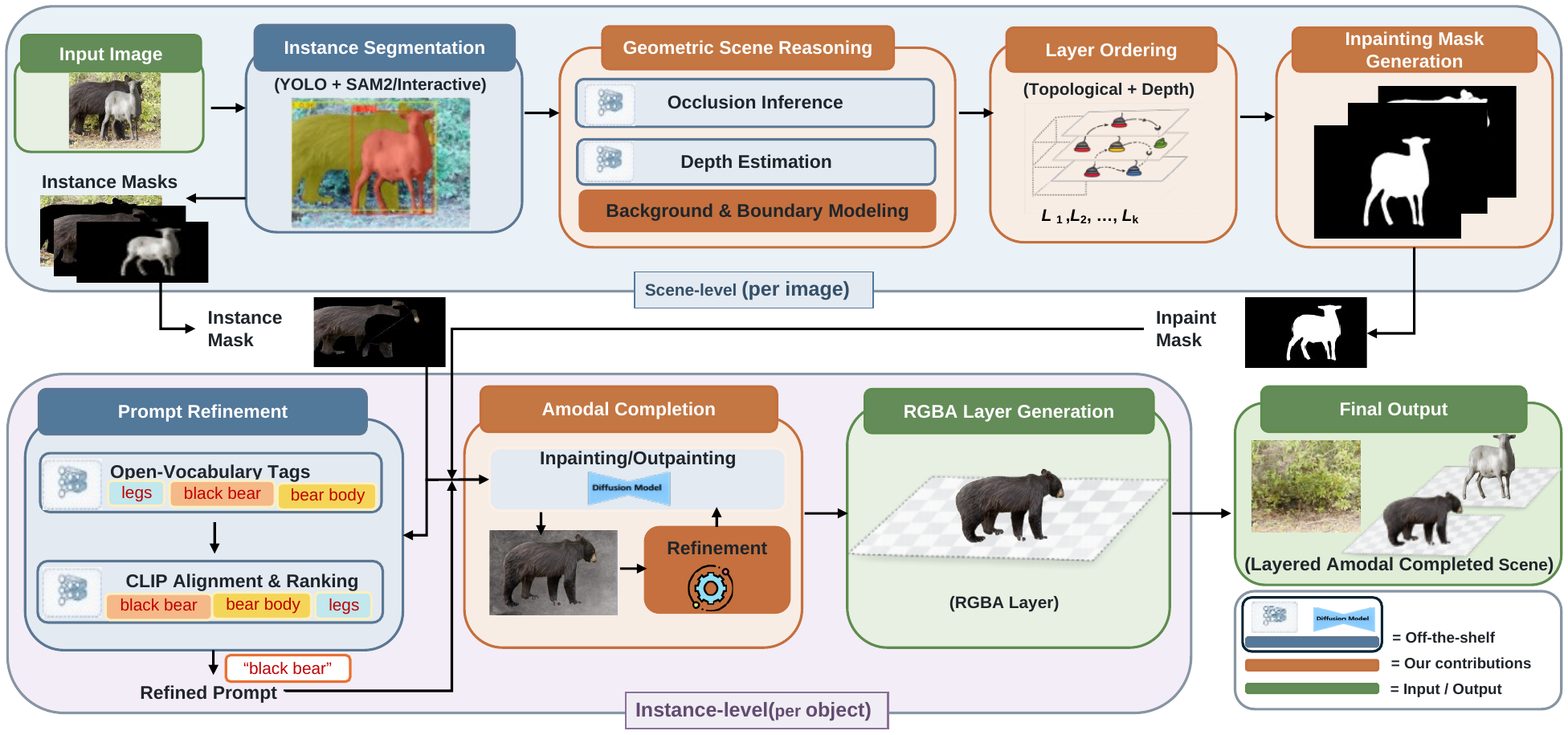}
    \caption[Layered amodal scene deocclusion framework]{%
    Overview of the layered amodal scene deocclusion framework. The system extracts object instances, performs geometric reasoning to construct an occlusion-aware scene representation, assigns depth-consistent layers, and generates structurally constrained masks together with refined prompts to guide diffusion-based amodal completion. The final output comprises compositable RGBA object layers that enable layered scene reconstruction. %
    }
    \label{fig:highlevel}
\end{figure*}

\section{Related Work}

\paragraph{Amodal Completion}
Amodal completion is the cognitive ability to perceive the full structure of an occluded object, long grounded in vision science through classical theories of shape continuity, symmetry, and perceptual familiarity \cite{Thielen2024AmodalCompletionBrain, restle1981classical}. Early computational approaches focused on semantic amodal segmentation, predicting binary masks for both visible and occluded regions \cite{Zhu2017SemanticAS, qi2019amodal}. These used a two-stage pipeline that first predicts the amodal mask and then completes it, but were limited to category-specific domains \cite{abs-1901-11390, engelcke2020genesis, abs-1903-00450, 10.5555/3495724.3497087, Yan2019VisualizingTI,10377684,10.1007/s11263-021-01517-0, Papadopoulos2019HowTM}. Self-supervised Scene Deocclusion (SSSD) \cite{zhan2020self} completed occluded objects by training on visible masks, but remained sensitive to segmentation errors and lacked image fidelity. Diffusion-based generative approaches improved completion realism. pix2gestalt \cite{ozguroglu2024pix2gestalt} trained a diffusion model to synthesize whole objects from partially visible counterparts, and Progressive Mixed Context Diffusion (PM-CD) \cite{xu2024amodal} completed objects by growing occluded regions progressively. Object-level Scene Deocclusion (PACO) \cite{10.1145/3641519.3657409} introduced a parallel visible-to-complete diffusion framework for scene-level completion, but faces challenges adapting to real-world environments. These methods are limited either in transferring from synthetic to real environments \cite{ozguroglu2024pix2gestalt, 10.1145/3641519.3657409} or to predefined categories \cite{xu2024amodal}. Recent work on real-world generalization completes arbitrary objects in naturalistic settings \cite{ao2025open}, but treats objects in isolation or relies on simplified background partitioning, which produces geometric inconsistencies in complex, multi-layered scenes. A related approach predicts a holistic occlusion-relation matrix in a single pass to handle mutual occlusion \cite{zhang2024hori}, but relies on a trained network and produces amodal masks rather than the training-free, appearance-level RGBA layer decomposition we target.

\paragraph{Vision-Language Models (VLMs)}

Current VLMs fall short on scene deocclusion. Models such as GPT-4o \cite{openai2024gpt4o} and Gemini \cite{gemini2023} struggle with basic spatial and occlusion reasoning, 
performing below human levels \cite{rahmanzadehgervi2024vision, pothiraj2025capture}. General-purpose VLMs like CogVLM \cite{CogVLM2024} and VisionLLM \cite{VisionLLM2024} excel at object recognition but are limited in occlusion reasoning, and LISA \cite{lai2023lisa} advances language-guided segmentation yet struggles with fully occluded regions. Text-conditioned editing models such as InstructPix2Pix \cite{brooks2022instructpix2pix} and OmniGen \cite{xiao2024omnigen} produce visually plausible outputs but lack any understanding of depth ordering or layered scene structure. Effectively addressing deocclusion requires explicitly modeling occlusion relationships and performing structured, layer-aware synthesis.

%% file: sec/3_method.tex
\section{Method}
We propose a unified, training-free framework for open-world layered amodal scene deocclusion (\cref{fig:highlevel}). The pipeline combines depth validated occlusion reasoning, cycle-safe layer assignment, and iterative diffusion refinement to produce globally consistent, physically grounded RGBA layers.

\subsection{Instance Detection and Segmentation}
\label{sec:instance_seg}
The pipeline supports two input modes, automatic and interactive, both producing binary instance masks in an identical format that is passed unchanged to downstream modules.

\noindent\textbf{Automatic Mode}
To prioritize boundary precision, we decouple detection from segmentation rather than relying on an end-to-end model, which tends to trade mask accuracy for recall. YOLO \cite{khanam2024yolov11overviewkeyarchitectural} produces bounding boxes $\mathcal{B}_{\text{auto}} = \{ (b_i, l_i, s_i) \}_{i=1}^{n}$ with class labels $l_i$ and confidence scores $s_i$ from the input image $I \in \mathbb{R}^{H \times W \times 3}$. Each box $b_i$ is then used as a prompt for SAM2 \cite{ravi2024sam2}, which returns a high-fidelity binary mask $M_i \in \{0,1\}^{H \times W}$. This yields instance masks $\mathcal{M}_{\text{auto}} = \{ M_1, \dots, M_n \}$ with provisional labels $\mathcal{L}_{\text{auto}} = \{ l_1, \dots, l_n \}$, while keeping detection and segmentation independently optimizable and avoiding error propagation along occlusion boundaries.

\noindent\textbf{Interactive Mode}
To operate beyond the YOLO vocabulary, users may specify arbitrary objects through point prompts $\mathcal{P} = \{ p_1, \dots, p_K \}$, each a 2D coordinate passed directly to SAM2 to produce a mask $M_k \in \{0,1\}^{H \times W}$. A default label is assigned unless the user provides one, giving $\mathcal{M}_{\text{inter}} = \{ M_{n+1}, \dots, M_{n+K} \}$ and $\mathcal{L}_{\text{inter}} = \{ \hat{l}_{n+1}, \dots, \hat{l}_{n+K} \}$.

Both modes share an identical mask format. The unified instance set entering all downstream stages is $\mathcal{M} = \mathcal{M}_{\text{auto}} \cup \mathcal{M}_{\text{inter}}$ with labels $\mathcal{L} = \mathcal{L}_{\text{auto}} \cup \mathcal{L}_{\text{inter}}$ and $N_{\text{total}} = |\mathcal{M}|$. All labels are provisional and refined at a later stage.

\subsection{Geometric Scene Reasoning via Occlusion and Depth Estimation}
\label{sec:scene_reasoning}
A geometrically faithful scene representation requires reasoning jointly about occlusion relationships, relative depth, and foreground structures that instance segmentation does not capture. Segmentation models detect semantically meaningful objects and, by design, do not segment every foreground surface in a scene \cite{He2017MaskR, DBLP:journals/corr/abs-2007-00047, Kirillov_2019_CVPR}. An occlusion graph built only from detected instances is therefore incomplete, since occlusions also arise from undetected foreground elements and from the image boundary itself. 
% Ignoring these sources yields a structurally deficient representation that limits accurate reasoning about which regions are occluded and the underlying causes.

\noindent\textbf{Pairwise Occlusion Relationship Inference.} For each spatially overlapping pair $(i,j)$, we infer the occlusion relationship using InstaOrder \cite{lee2022instaorder}, represented as a binary adjacency matrix $\mathcal{O} \in \{0,1\}^{N \times N}$ where $\mathcal{O}_{ij}=1$ indicates that object $i$ occludes object $j$. This matrix is the primary structural prior.

\noindent\textbf{Dense Monocular Depth Estimation.}
\label{mono_depth}Pairwise occlusion gives ordinal front/back relations but not continuous structure such as intra-object depth variation or fine transitions near occlusion boundaries. We recover this signal with Depth Anything V2 \cite{depth_anything_v2}, producing a dense relative depth map $D \in \mathbb{R}^{H \times W}$ in which larger values are closer to the camera. Each object $i$ is summarized by its median masked depth $d_i = \mathrm{median}\{D(p) \mid p \in M_i\}$, a more geometrically reliable per-object estimate than the mean.

\noindent\textbf{Depth-driven background occluder detection.}
\label{background_occluder}
%Given the union of instance masks $\mathcal{M}$, the background is its complement $\bar{\mathcal{M}} = 1 - \mathcal{M}$.
The background is the complement of the pixel-wise union of all instance masks, $\bar{M} = 1 - \bigvee_{i=1}^{N_{\text{total}}} M_i$. This background may include undetected foreground occluders, so we now test for that. For each object $i$ with mask $M_i$ and bounding box $b_i$, we examine the local background region $\mathcal{R}_i = \bar{M} \cap b_i$, and classify a pixel $p \in \mathcal{R}_i$ as an occluder candidate when its depth exceeds that of object $i$ by a proportional margin $\tau$, as defined in \cref{eq:occluder_depth}:
\begin{equation}
D(p) > d_i\,(1 + \tau)
\label{eq:occluder_depth}
\end{equation}
The multiplicative margin scales the threshold with object depth, giving consistent behavior across the relative, unitless values from monocular estimation. Candidates with negligible signal are discarded, and the remainder pass a proximity filter that retains only connected components adjacent to the object boundary, removing background regions that share a similar depth but are geometrically unrelated to the local occlusion. Accepted candidates are split into connected components and lightly dilated to close boundary gaps.

In contrast, morphological approaches partition background pixels purely by spatial adjacency, with no awareness of scene geometry, and recompute the partition independently per query object \cite{ao2025open}. This cannot distinguish physically distinct background surfaces that share a boundary, yielding a single ambiguous decision per query. We instead group fragments \emph{globally} across all scene entities using geometric cues. Fragments form a similarity graph $G_f = (\{f_j\}, E_f)$, with an edge $(f_j, f_k)$ whenever the similarity $s(f_j, f_k)$ in \cref{eq:similarity} exceeds a threshold:
\begin{equation}
s(f_j, f_k) = w_d\, s_{\text{depth}}\, s_{\text{spatial}} + w_c\, s_{\text{color}} + w_t\, s_{\text{texture}}(f_j, f_k)
\label{eq:similarity}
\end{equation}
Multiplying $s_{\text{depth}}$ and $s_{\text{spatial}}$ rather than adding them encodes a geometric prior: spatial proximity is a necessary condition for grouping, so however similar two fragments are in depth or appearance, merging them across a large spatial gap remains implausible. This places fragments from physically distinct surfaces into separate groups before any occlusion decision. Given a confirmed group, we reconstruct the complete occluder mask via SAM based shape completion, falling back to a depth guided convex hull if SAM fails, and validate each candidate against overlap, size, and depth consistency. The completed occluder masks are inserted as dedicated nodes in the occlusion graph, extending it beyond the object-only convention of InstaOrder \cite{lee2022instaorder} to capture foreground occlusion sources that segmentation alone cannot recover.

\noindent\textbf{Boundary-aware occlusion detection.} For each instance $j$, we test whether the contour of $M_j$ touches the image boundary. Let $\mathcal{E} \subseteq \{\text{top}, \text{bottom}, \text{left}, \text{right}\}$ be the set of contacted edges. If $\mathcal{E} \neq \emptyset$, the boundary acts as a truncation source whose missing content lies outside the frame. For each $e \in \mathcal{E}$ we add a virtual truncation node $\mathcal{V}_e$ with a directed edge $\mathcal{V}_e \rightarrow j$. The node carries no pixel support, as truncation is structural rather than spatial, and the identity of $e$ encodes the completion direction. This flags boundary-truncated instances for directed shape completion instead of treating them as fully visible.

\subsection{Topological-depth Layer Assignment and Inpainting Mask Generation}
\label{sec:layering}

\noindent\textbf{Topological-depth Layer Assignment.}
Given $N_\text{vis}$ instances with masks $\{ M_i \}_{i=1}^{N_\text{vis}}$, a depth map $D \in \mathbb{R}^{H \times W}$, and the occlusion matrix $\mathcal{O} \in \{0,1\}^{N_\text{vis} \times N_\text{vis}}$, we assign each object a discrete layer $\ell_i \in \mathbb{Z}_{\ge 1}$ such that occluders receive higher layers (closer to the camera) than the objects they occlude. Monocular depth is unreliable near boundaries and discontinuities, whereas the occlusion graph gives reliable pairwise order but cannot resolve relative depth among non-adjacent or peer objects. We therefore combine the two in two stages.

\noindent\textbf{Stage 1: Longest-path grouping.} Each object $i$ is scored by the longest occlusion chain it dominates: $g_i = 1$ if it occludes nothing, else $g_i = 1 + \max_{j:O_{ij}=1} g_j$. Grouping by $G_k = \{i \mid g_i = k\}$ places every occluder in a strictly higher group than any object it occludes ($g_i \geq g_j + p$ along any path of length $p$). This recursion is well-defined only on acyclic graphs, so mutual-occlusion cycles are first contracted into single nodes via strongly connected components, becoming depth-peer sets whose internal order defers to Stage~2. The result is a coarse layering that uses only discrete occlusion and is immune to depth noise.

\noindent\textbf{Stage 2: Depth-based separation.} Within each group $G_k$, we resolve the remaining ambiguity using the median depth $d_i$ (\cref{mono_depth}). Since the depth model \cite{depth_anything_v2} follows the larger-is-closer convention, objects within a group are ordered by ascending $d_i$ (back to front), with closer objects assigned higher layers. This settles the order within each group, including peers inside a cycle that occlusion alone leaves tied, while preserving the coarse inter-group order established by discrete occlusion, which is unaffected by depth noise.

\noindent\textbf{Inpainting Mask Generation.} For each target object $i$ we
construct a binary inpainting mask $\mathcal{I}_i \in \{0,1\}^{H \times W}$ marking the region to complete. The transitive closure finds all objects that directly or indirectly occlude the target, and the initial mask $\mathcal{I}_i^{(0)}$ is the union of their contributions: the full mask for ordinary occluders, but only the shared frontier region for mutually occluding objects, avoiding painting their entire silhouette into the target. A final depth-guided reachability step then constrains this union: the occluder region within the target's bounding box is always retained as a minimal region $\mathcal{I}_i^{\text{min}}$, and beyond it a pixel is kept only if it is spatially connected to this core and consistent with a local depth reference, avoiding the unreliability of a single global depth near occlusion boundaries. The final mask stays within the valid occluders, $\mathcal{I}_i^{\text{min}} \subseteq \mathcal{I}_i^{\text{final}} \subseteq \mathcal{I}_i^{(0)}$.

\subsection{Prompt Refinement and Amodal Completion}
\label{sec:prompt_refine_amodal_comp}
Each object layer is paired with its inpainting mask $\mathcal{I}_i$, whose marked region is completed by a diffusion model conditioned on an object-specific text prompt refined in two stages.

\noindent\textbf{Prompt Refinement.} Coarse detector labels often miss fine-grained appearance, causing semantic drift in the completion. We refine each prompt by open-vocabulary tagging followed by cross-modal selection. First, the masked object is composited onto a white background to suppress bias from surrounding regions and passed to Recognize-Anything Plus \cite{huang2023open}, yielding semantic tags $\mathcal{T}_{\mathrm{RAM}} = \{t_1, t_2, \ldots\}$. The candidate set $\mathcal{C} = \{l_{\text{det}}\} \cup \mathcal{T}_{\text{RAM}}$ combines these with the detector label. Each candidate and the masked image are encoded with CLIP \cite{pmlr-v139-radford21a}, and the label maximizing image–text cosine similarity is selected as the grounded prompt for diffusion.

\noindent\textbf{Amodal Completion.} Objects fully within the image but partially occluded are completed by conditional inpainting; boundary-truncated objects by conditional outpainting; objects that are both receive inpainting of the interior occlusion, then outpainting to extend the truncated support.

For \emph{inpainting}, the binary mask $\mathcal{I}_i$ is feathered into a soft weighting field $\tilde{\mathcal{I}}_i \in [0,1]^{H \times W}$ that smooths the transition between visible and inferred regions. The layer image, soft mask, and refined prompt are passed to a pretrained latent diffusion inpainter \cite{Rombach_2022_CVPR}. Motivated by \cite{xu2024amodal}, we condition completion on the target's own visible region rather than the surrounding scene, excluding other objects and background to avoid co-occurrence bias. The synthesized region is then blended back with mild boundary smoothing.

For \emph{outpainting}, the canvas is expanded in empirically chosen fixed increments of $\Delta_{\text{out}} = 150$ pixels along each truncated edge until the predicted amodal extent no longer reaches the new boundary or a cap of three expansions is hit, so the total extension follows the object's geometry rather than a fixed scale. Expanded regions are initialized with a neutral background to avoid prior bias \cite{xu2024amodal}, and the outpainting mask covers only the new area with a slight overlap into the original. A protection mask $\alpha_{\text{obj}} \in [0,1]^{H \times W}$ over known object pixels keeps the visible content unchanged, $C = \alpha_{\text{obj}}\, I + (1 - \alpha_{\text{obj}})\, I_{\text{gen}}$, with $I_{\text{gen}}$ the generated content.

\begin{figure}[t]
    \centering
    \includegraphics[width=\linewidth]{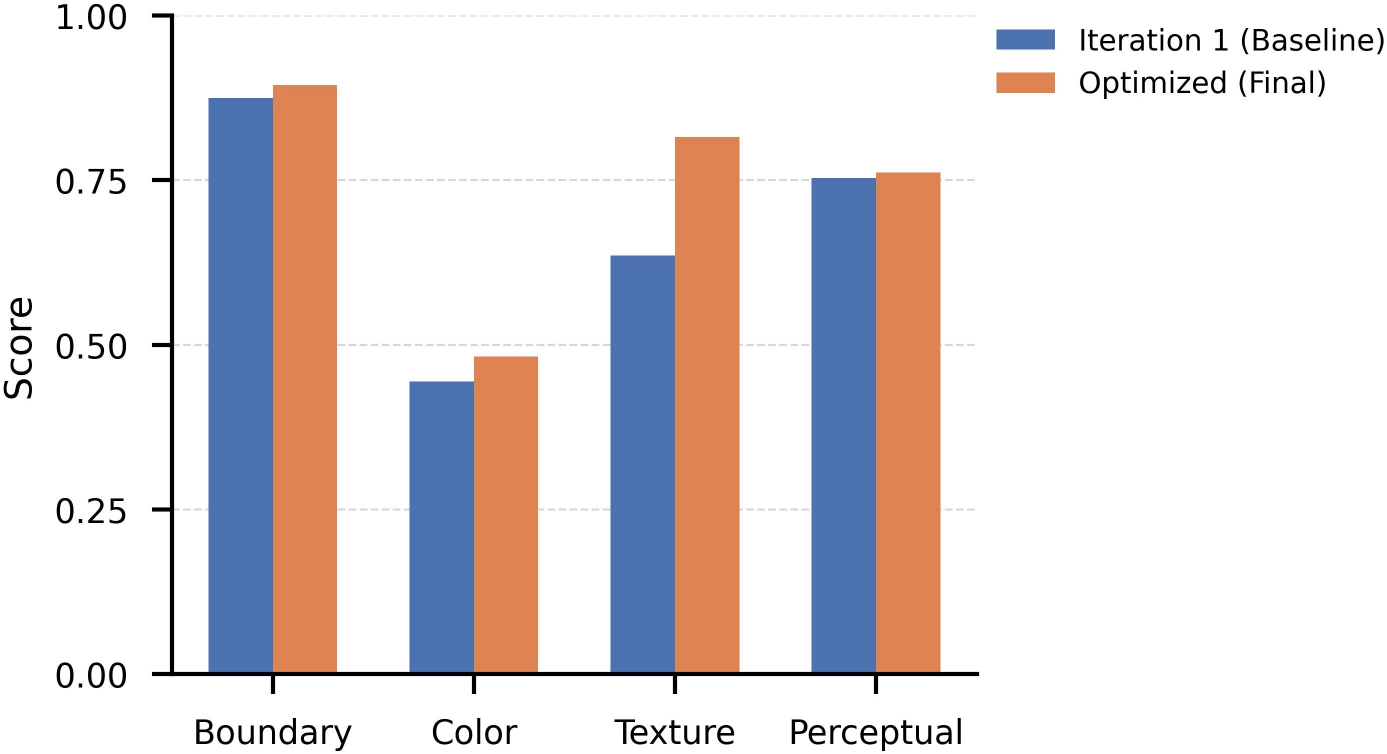}
    \caption{Optimization impact across the four quality dimensions ($s_1$--$s_4$) between the baseline (iteration~1) and the final output.}
    \label{fig:metric_bar}
\end{figure}

\noindent\textbf{Quality-Driven Iterative Refinement.} To ensure structural and perceptual integrity, completions pass through an iterative loop guided by a composite quality score $\mathcal{Q}$ that aggregates four equally weighted criteria $s_k \in [0,1]$ under a per-component constraint, as defined in \cref{eq:quality}:
\begin{equation}
\mathcal{Q} = \frac{1}{4}\sum_{k=1}^{4} s_k, \quad \text{s.t.} \quad s_k \geq \tau_k .
\label{eq:quality}
\end{equation}
For each object layer in isolation, the criteria assess complementary dimensions: boundary continuity ($s_1$) from local gradient alignment across the completion interface, color fidelity ($s_2$) from CIE $\Delta E$ transitions in LAB space, texture coherence ($s_3$) from edge-density ratios and coefficients of variation, and global plausibility ($s_4$) from luminance variance and color diversity. Because these are low-level image statistics, each failure mode maps directly to a score component. Thresholds adapt to scene complexity through a baseline $\mathcal{Q}_{\text{base}}$, the mean of the four sub-scores on the visible region, giving $\tau_k = \alpha_k\ \cdot \mathcal{Q}_{\text{base}}$ with $\alpha_k$ set by the structural difficulty of the occlusion.

The loop terminates only when both global quality and every component constraint are met, avoiding over-smoothing or over-sharpening. Of the four, color fidelity ($s_2$) is the most resistant to iterative gain, while texture coherence ($s_3$) improves most (\cref{fig:metric_bar}); the other two are typically strong from the first iteration. Generation is steered by the weakest sub-score, exploring at most $N{=}6$ iterations over $C{=}8$ predefined configurations that vary denoising steps, guidance scale, feathering, and mask dilation (\cref{fig:ConfigBenchmarking}). This raises the completion pass rate from $12.49\%$ to $35.59\%$ ($2.85\times$), with $97.23\%$ of recoverable layers resolved within six iterations.

\noindent\textbf{RGBA Output.} After completion, each object is exported as an RGBA image $I_{\text{comp}} \in \mathbb{R}^{H \times W \times 4}$, where the RGB channels hold the blended appearance and the alpha channel encodes the amodal object mask, giving a compositable per-object layer.

%% file: sec/4_experiments.tex
\begin{figure*}[!t]
    \centering
    \includegraphics[width=\textwidth]{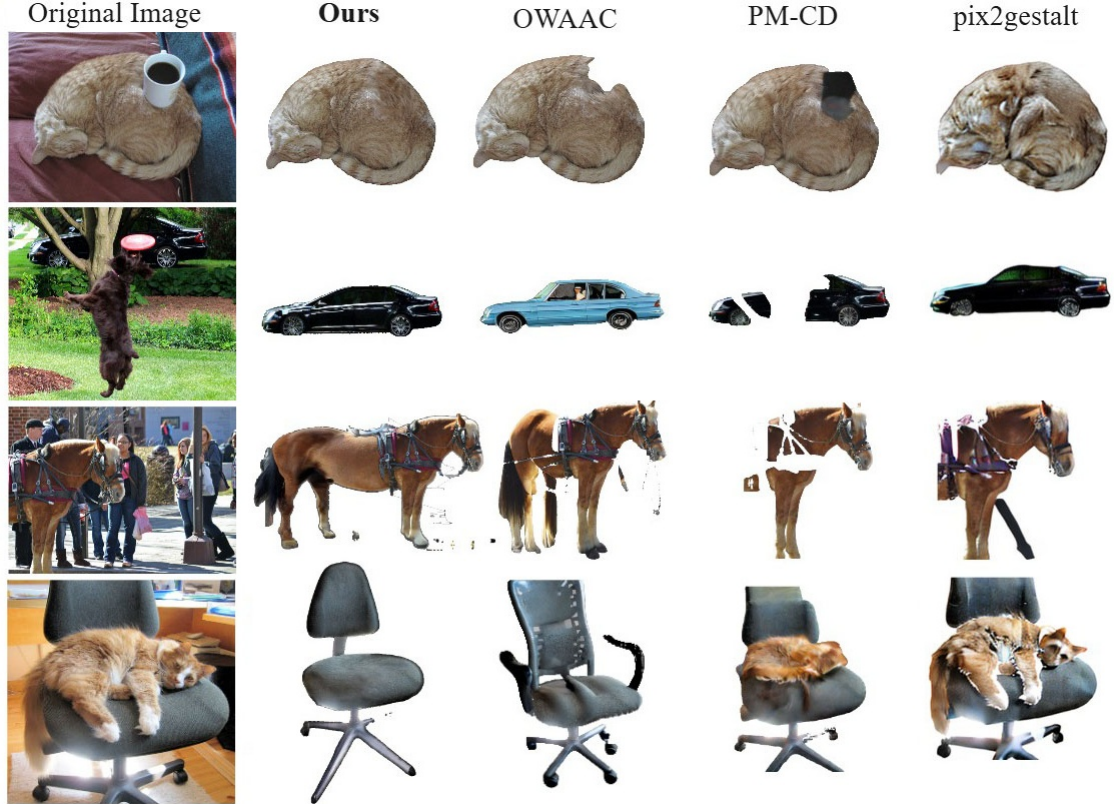}
    \caption{Qualitative comparison with existing approaches. Our method shows improved structural consistency and perceptual quality across diverse occlusion patterns. OWAAC produces objects that are semantically similar but visually different (car, chair), PM-CD leaves large gaps in objects (car, horse), pix2gestalt does the same to a lesser extent (horse, cat on chair).}
    \label{fig:comparison}
\end{figure*}
\section{Experiments}
\subsection{Datasets}
\noindent\textbf{Pseudo-Occluded Dataset.} Natural images with reliable amodal ground truth are scarce. We therefore build a controlled pseudo-occluded set from COCO \cite{DBLP:journals/corr/LinMBHPRDZ14}: 2{,}745 images across roughly 70 categories, where a second object is composited over a fully visible target at controlled positions. The pre-composition annotation serves as unambiguous amodal ground truth, and occlusion severity is varied from 20\% to 90\%.

\noindent\textbf{Real-World Datasets.} To assess generalization, we evaluate on real occlusions from COCO-A \cite{Zhu2017SemanticAS}, Visual Genome (VG) \cite{krishna2017visualgenome}, and copyright-free images previously used by OWAAC \cite{ao2025open}. As these lack amodal annotations, we compare predicted modal and amodal regions within the inferred completion area. This proxy does not measure geometric amodal accuracy directly, but quantifies perceptual fidelity and semantic consistency in the hallucinated regions, and follows the same evaluation protocol adopted by OWAAC \cite{ao2025open} for this task.

\subsection{Comparison With Other Methods}
We compare against three recent methods, OWAAC \cite{ao2025open}, PM-CD \cite{xu2024amodal}, and pix2gestalt \cite{ozguroglu2024pix2gestalt}, both quantitatively (\cref{tab:quant_results}) and qualitatively (\cref{fig:comparison}). We were unable to include PACO \cite{10.1145/3641519.3657409} as a baseline, as no runnable public implementation was available at the time of writing. We report metrics at three perceptual levels: CLIP score \cite{pmlr-v139-radford21a} for high-level semantic consistency (cosine similarity of image embeddings), LPIPS \cite{Zhang2018TheUE} and VGG16 feature similarity \cite{Gatys2016ImageST} at the mid level, and SSIM for low-level structural fidelity. As shown in \cref{tab:quant_results}, our method outperforms prior work across all four datasets.

Because amodal completion is inherently subjective, we also ran a perceptual user study. Twenty-seven participants each evaluated 25 randomly sampled completions across four methods (ours, pix2gestalt, PM-CD, OWAAC), giving 675 ratings per method. Each was shown the original image with the four completions in randomized order, and used two protocols: a forced-choice selection of the single most realistic completion, and an independent 5-point Likert quality rating. This decouples relative preference from absolute quality. Our method was preferred in 68.0\% of forced choice trials, against PM-CD (12.1\%), OWAAC (10.7\%), and pix2gestalt (9.2\%). Overall, our completions received a quality rating of between~3 and~5 (Acceptable, Good, or
Excellent) in 83.6\% of cases, compared with 33.1\% (OWAAC), 29.2\% (PM-CD), and 29.9\% (pix2gestalt), confirming consistent perceptual quality.

\begin{table}[t]
\caption{Quantitative evaluation against state-of-the-art methods across perceptual levels. $\uparrow$/$\downarrow$ indicate higher/lower is better; best in \textbf{bold}.}
\label{tab:quant_results}
\centering
\scriptsize
\setlength{\tabcolsep}{4pt}
\renewcommand{\arraystretch}{1.2}
\begin{tabular}{llcccc}
\hline
Dataset & Method & CLIP $\uparrow$ & LPIPS $\downarrow$ & Feat. Sim. $\uparrow$ & SSIM $\uparrow$ \\
\hline
\textbf{Pseudo-occluded}
& pix2gestalt & 26.462 & 0.427 & 0.513 & 0.643 \\
& PM-CD       & 26.947 & 0.638 & 0.337 & 0.412 \\
& OWAAC       & 27.105 & 0.384 & 0.598 & 0.726 \\
& Ours        & \textbf{28.241} & \textbf{0.230} & \textbf{0.798} & \textbf{0.895} \\
\hline
\textbf{COCO-A}
& pix2gestalt & 26.998 & 0.471 & 0.524 & 0.695 \\
& PM-CD       & 27.383 & 0.664 & 0.328 & 0.382 \\
& OWAAC       & 27.612 & 0.351 & 0.609 & 0.718 \\
& Ours        & \textbf{27.826} & \textbf{0.241} & \textbf{0.771} & \textbf{0.859} \\
\hline
\textbf{VG}
& pix2gestalt & 27.672 & 0.429 & 0.554 & 0.726 \\
& PM-CD       & 28.367 & 0.578 & 0.413 & 0.463 \\
& OWAAC       & 28.470 & 0.310 & 0.658 & 0.732 \\
& Ours        & \textbf{28.974} & \textbf{0.201} & \textbf{0.816} & \textbf{0.891} \\
\hline
\textbf{Free Images}
& pix2gestalt & 27.621 & 0.393 & 0.613 & 0.732 \\
& PM-CD       & 28.333 & 0.720 & 0.279 & 0.309 \\
& OWAAC       & 28.652 & 0.269 & 0.698 & 0.753 \\
& Ours        & \textbf{28.870} & \textbf{0.243} & \textbf{0.836} & \textbf{0.814} \\
\hline
\end{tabular}
\end{table}

\subsection{Ablation Studies}
We ablate each novel component on the combined COCO-A and VG split (\cref{tab:ablation}). Removing cycle handling (naïve topological sorting with no mutual-occlusion masking), background occluder detection, or iterative refinement degrades every metric, confirming each contributes to the final result. Off-the-shelf components (InstaOrder, Depth Anything V2, SAM2) are used unmodified and are not ablated.

\begin{table}[t]
\caption{Ablation of each contribution, evaluated on the combined COCO-A and VG split.}
\label{tab:ablation}
\centering
\scriptsize
\setlength{\tabcolsep}{6pt}
\renewcommand{\arraystretch}{1.4}
\begin{tabular}{lcccc}
\hline
Configuration & CLIP $\uparrow$ & LPIPS $\downarrow$ & Feat. $\uparrow$ & SSIM $\uparrow$ \\
\hline
w/o cycle handling    & 28.323 & 0.221 & 0.785 & 0.862 \\
w/o background occluder det.  & 28.258 & 0.237 & 0.772 & 0.854 \\
w/o iterative refinement      & 28.204 & 0.213 & 0.802 & 0.883 \\
Full system                   & \textbf{28.923} & \textbf{0.203} & \textbf{0.814} & \textbf{0.890} \\
\hline
\end{tabular}
\end{table}

\begin{figure}[t]
    \centering
    \includegraphics[width=\linewidth]{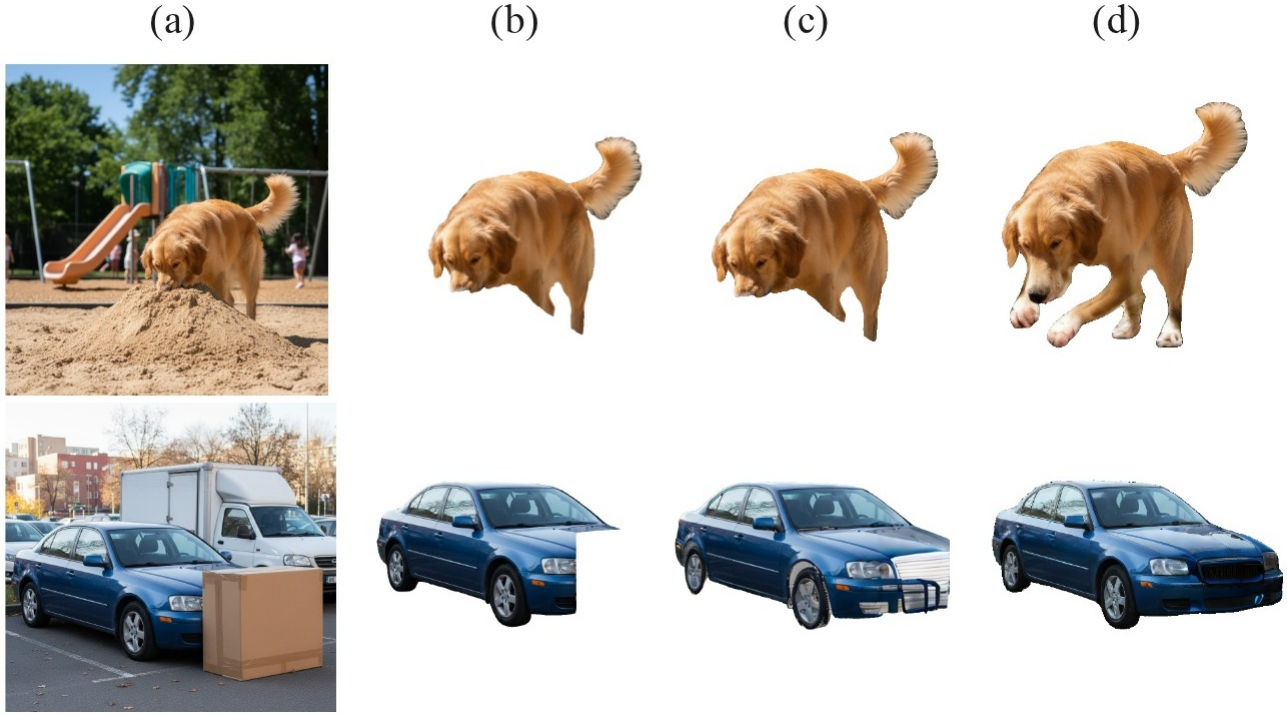}
    \caption{Ablation on background occluder detection. (a) Input. (b) Without detection, the dog's legs and the car's front are not recovered. (c) The morphological baseline cannot separate geometrically distinct background surfaces from the foreground, giving inaccurate completions. (d) Our depth-guided approach isolates each background occluder for accurate completion.}
    \label{fig:BOAblation}
\end{figure}

\noindent\textbf{Background occluder detection.} We compare three configurations: without background occluder detection, morphological-only detection, and our full depth-guided pipeline (\cref{background_occluder}). Without detection, occlusions from background surfaces are never resolved, as the occluding region is never recovered. The morphological baseline recovers some boundaries but fails where geometrically distinct background surfaces are spatially adjacent, grouping them by proximity alone and losing the background as a distinct surface. \cref{fig:BOAblation} shows the three configurations under background occlusion.

We quantify our depth-guided background occluder detector and the morphological baseline of \cite{ao2025open} on $100$ manually annotated COCO images \cite{DBLP:journals/corr/LinMBHPRDZ14}: $70$ occluder-present images for detection and localization (\cref{tab:bg_occluder}), and $30$ occluder-free images to probe false positives directly. At the operative $0.25$ threshold, our detector recovers $66$ of $70$ occluders with no false positives on either set. The four missed occluders are thin, low-signal structures that fall below the acceptance margin $\tau$, exactly the regime the candidate filter is designed to discard. The morphological baseline misses most occluders and falsely flags adjacent background surfaces ($29$ false positives on occluder-present cases and $9$ on negatives). Since our detections are false-positive-free on this benchmark, the IoU sweep in \cref{tab:bg_occluder} isolates localization, where our margin remains clear across thresholds.The detected mask only seeds $\mathcal{I}_i^{(0)}$ before dilation and refinement (\cref{sec:layering}), so approximate localization suffices for reliable completion.

\begin{table}[t]
\caption{Background occluder detection on $70$ occluder-present COCO
images~\cite{DBLP:journals/corr/LinMBHPRDZ14}, comparing our depth-guided detector with a morphological baseline across IoU thresholds. Each detector predicts a fixed set of occluders per image; only the IoU matching changes across rows. Brackets on the $0.25$ row are bootstrap $95\%$ CIs; with no false positives, ours reports a rule-of-three precision bound ($\ge 0.955$). Best in \textbf{bold}.}
\label{tab:bg_occluder}
\centering
\scriptsize
\setlength{\tabcolsep}{4pt}
\renewcommand{\arraystretch}{1.4}
\begin{tabular}{llccc}
\hline
Method & IoU & Precision $\uparrow$ & Recall $\uparrow$ & F1 $\uparrow$ \\
\hline
\textbf{Morph.}~\cite{ao2025open}
  & 0.25 & 0.216 [0.091, 0.355] & 0.114 [0.043, 0.200] & 0.150 \\
  & 0.50 & 0.054 & 0.029 & 0.037 \\
  & 0.75 & 0.027 & 0.014 & 0.019 \\
\hline
\textbf{Ours}
  & 0.25 & \textbf{1.000} ($\ge$0.955) & \textbf{0.943} [0.886, 0.986] & \textbf{0.971} \\
  & 0.50 & \textbf{0.818} & \textbf{0.771} & \textbf{0.794} \\
  & 0.75 & \textbf{0.485} & \textbf{0.457} & \textbf{0.471} \\
\hline
\end{tabular}
\end{table}

\begin{figure}[htp]
    \centering
    \includegraphics[width=\linewidth]{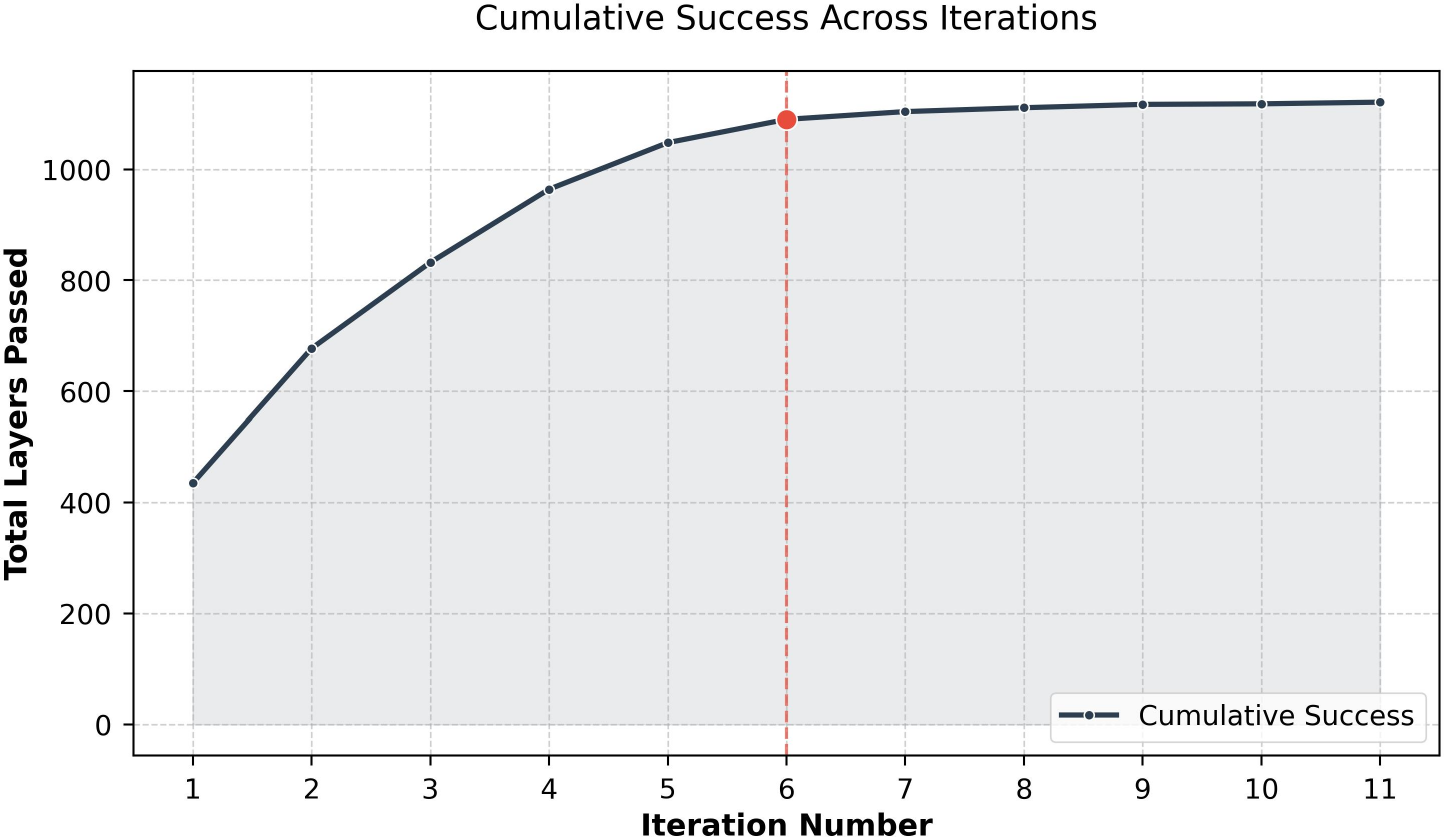}
    \caption{Ablation analysis of the iterative refinement pipeline: Cumulative layers passing the quality threshold across iterations; the red dashed line at $N{=}6$ marks the point of diminishing returns. }
    \label{fig:CumulativeSuccess}
\end{figure}

\begin{figure}[htp]
    \centering
    \includegraphics[width=\linewidth]{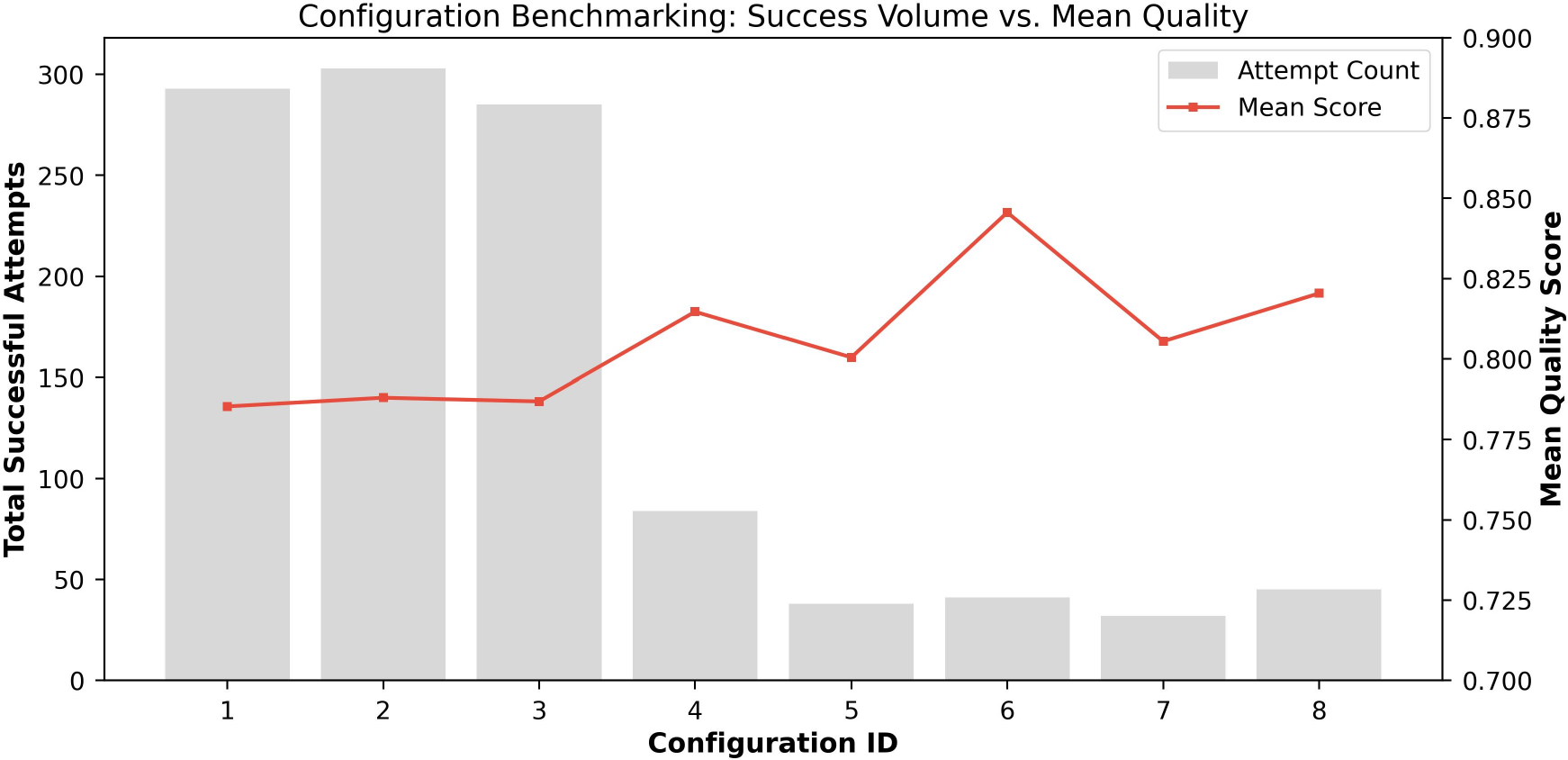}
    \caption{Ablation analysis of the iterative refinement pipeline: Success volume (bars) and mean quality score (line) per configuration, 
suggesting a trade-off between selection frequency and 
output quality across the $C{=}8$ parameter configurations. }
    \label{fig:ConfigBenchmarking}
\end{figure}

\noindent\textbf{Iterative refinement.} \cref{fig:CumulativeSuccess} shows cumulative pass rate across iterations, rising sharply through $N{=}6$ to recover $97.23\%$ of achievable successes before gains flatten; we adopt $N{=}6$ as the quality--cost trade-off. \cref{fig:ConfigBenchmarking} shows the trade-off across the $C{=}8$ configurations: configurations 1--3 account for $78.6\%$ of successes through broad applicability, while configuration 6 reaches the highest mean quality ($\mathcal{Q}{=}0.846$) at low coverage ($3.66\%$). This indicates that failure-mode-driven selection across configurations balances quality and pass rate.


\noindent\textbf{Efficiency.} All experiments run on a single NVIDIA RTX A4000 (16\,GB). A single object image takes about 28\,s end to end: image level stages (detection, segmentation, geometric reasoning, layering) take $\approx$3.1\,s combined, and per-object prompt refinement and completion account for the rest, scaling linearly with object count. Peak GPU memory is 9--13.5\,GB during completion and under 9\,GB for all upstream stages.

%% file: sec/5_conclusion.tex
\section{Conclusion}
We presented an open-world, training-free framework for scene deocclusion that recovers partially occluded and boundary-truncated objects within a layered representation. Built on pretrained foundation models and depth-guided geometric reasoning, it resolves both undetected occluders and cyclic occlusions, and produces perceptually coherent completions that outperform recent state-of-the-art methods across four datasets and a user study. The resulting RGBA layers form a compositable representation that integrates directly into downstream tasks such as image editing, 3D reconstruction \cite{Tudosiu_2024_CVPR, szymanowicz24splatter}, novel view synthesis, and other vision pipelines \cite{gkitsas2021panodr, 10.5555/3495724.3497087, 10.1007/978-3-031-19784-0_41, Pintore2022InstantAE}.

\noindent\textbf{Limitations.} Stacking pretrained models makes the pipeline susceptible to error propagation, although its modular design allows individual components to be upgraded independently. With limited visible evidence, diffusion may produce plausible but globally inconsistent completions; resampling helps but cannot replace missing observations. Incorporating geometric priors and cross-layer context is a promising direction for improving this consistency. Thin or low-signal occluders may also fall below the detection threshold and remain unrecovered. Finally, for mutually occluding objects, a single 2D layer stack cannot preserve the
cyclic relation once the pair is separated during editing, so faithful cyclic
recomposition requires relationship-aware reasoning beyond a layer stack, which we leave to future work.

% For genuine occlusion cycles, our depth-based ordering yields a consistent layering that is sufficient for de-occlusion, since each object's completion draws on its occluders regardless of their relative ordering. It is necessarily approximate for 2D recomposition. However, no single linear order can preserve every pairwise relation in a cycle, so faithfully editing mutually occluding objects requires joint, relationship-aware reasoning beyond a layer stack. Incorporating geometric priors and cross-layer context is therefore a promising direction for future work.